\documentclass{article} % For LaTeX2e
\usepackage{iclr2022_conference,times}

\usepackage{amsmath,amsfonts,bm}

\def\eqref#1{equation~\ref{#1}}
\def\1{\bm{1}}

\DeclareMathAlphabet{\mathsfit}{\encodingdefault}{\sfdefault}{m}{sl}
\SetMathAlphabet{\mathsfit}{bold}{\encodingdefault}{\sfdefault}{bx}{n}

\usepackage{hyperref}
\usepackage{url}
\usepackage{multirow}
\usepackage[pdftex]{graphicx}
\usepackage{hhline}
\usepackage{makecell}
\usepackage[utf8x]{inputenc}

\title{miniF2F: a cross-system benchmark for \\formal Olympiad-level mathematics}

\author{Kunhao Zheng \\
\'Ecole Polytechnique\\
\texttt{kunhao.zheng@polytechnique.edu} \\
\And
Jesse Michael Han\\
OpenAI \\
University of Pittsburgh \\
\texttt{jessemichaelhan@openai.com} \\
\AND
Stanislas Polu \\
OpenAI \\
\texttt{spolu@openai.com}
}

\iclrfinalcopy % Uncomment for camera-ready version, but NOT for submission.
\begin{document}

\maketitle

\begin{abstract}
We present $\textsf{miniF2F}$, a dataset of formal Olympiad-level mathematics problems statements intended to provide a unified cross-system benchmark for neural theorem proving. The \textsf{miniF2F} benchmark currently targets Metamath, Lean, Isabelle (partially) and HOL Light (partially)
and consists of 488 problem statements drawn from the AIME, AMC, and the International Mathematical Olympiad (IMO), as well as material from high-school and undergraduate mathematics courses. We report baseline results using GPT-$f$~\citep{polu2020generative}, a neural theorem prover based on GPT-3~\citep{brown2020language} and provide an analysis of its performance. We intend for \textsf{miniF2F} to be a community-driven effort and hope that our benchmark will help spur advances in neural theorem proving.
\end{abstract}

\section{Introduction}
% TODO(kunhao/jesse): https://www.cs.ru.nl/~freek/100/

% notes on content of introduction
% compare to the Freek 100 benchmark, emphasize how cross-platform benchmarks have spurred the development of proof assistant libraries
% before this, bring up the central role of shared benchmarks and datasets within deep learning (e.g. most famously ImageNet). Introduce MiniF2F in the 2nd or 3rd paragraph, explaining its motivation wrt the IMO grand challenge.
% remove mention of deep reinforcement learning
% add a sentence which acknowledges automated theorem proving benchmarks
% intended role of miniF2F is to provide a common resource to research groups working on formal theorem proving 
% scientific contribution: baseline results using large language models
% also emphasize that large language models are not a mandatory approach for Olympiad problems and they are not assumed in any way within the dataset --- motivation for this is to emphasize the generality and widespread applicability of the dataset
% unique selling point: (1) formal, i.e. end-to-end verifiable and not just for concrete cases, (2) cross-platform, and (3) specific, well-defined scope: olympiad level problems
% scientific contribution: report baseline results
% possible TODO: report basic expert iteration results on miniF2F-valid and miniF2F-test, without mentioning investigation into the curriculum hypothesis.

% ===========================
% ======= begin of v2 =======
Shared benchmarks and datasets have historically played a crucial role in driving advances in large-scale applications of deep learning, e.g. in computer vision \citep{deng2009imagenet} and natural language processing
\citep{DBLP:conf/iclr/WangSMHLB19, DBLP:conf/emnlp/RajpurkarZLL16, DBLP:conf/acl/PapernoKLPBPBBF16}. \emph{Neural theorem proving} is a rapidly developing area which aims to apply techniques from deep learning to interactive theorem proving. To date, most contributions in this area have focused on individual theorem proving systems, each with a separately-implemented mathematics library and with results reported on a dataset-specific test split; examples include the HOList~\citep{bansal2019holist}, CoqGym~\citep{yang2019learning} and LeanStep~\citep{han2021proof} theorem proving environments and benchmarks. However, benchmarks from this paradigm are not ideal for measuring the mathematical reasoning ability of neural theorem provers for several reasons. Library-specific train/test splits are siloed by construction, dependent on how theorems and lemmas are split in these libraries, and as such are not directly comparable across systems. Moreover, formal mathematics libraries are closer to software repositories than informal mathematical exposition, and many lemmas are implementation-specific artifacts without precise informal mathematical or cross-system translations.

To date, the neural theorem proving community has not organized its efforts around a cross-system benchmark. To address this need and to provide a common resource to research groups working on formal theorem proving, we present \textsf{miniF2F}, a unified cross-system benchmark of formal mathematics of progressively increasing difficulty, centering around Olympiad-level problem statements (AMC, AIME, IMO) as well as high-school and undergraduate maths classes. Both the content and name of \textsf{miniF2F} are inspired by the IMO Grand Challenge~\citep{imograndchallenge}: to build an AI that can win a gold medal in the International Mathematical Olympiad in a \emph{formal-to-formal} (F2F) format. More precisely, the agent must receive IMO problems written in a formal mathematical format, and must produce a formal (i.e. machine-checkable) proof for that problem.

We intend for \textsf{miniF2F} to serve as a stepping stone for different formal systems towards the IMO Grand Challenge~\citep{imograndchallenge}, as it is end-to-end verifiable, cross-platform and spans a wide range of difficulty. While we report baseline results on \textsf{miniF2F} using GPT-$f$, a language model based on GPT-3 which has been finetuned for theorem proving, language models are not a mandatory approach for Olympiad problems and this assumption is not reflected in \textsf{miniF2F}, preserving the generality and widespread applicability of the benchmark to systems similar to DeepHOL~\citep{bansal2019holist} or Holophrasm~\citep{DBLP:journals/corr/Whalen16}.

\section{Background and related work}
% TODO(kunhao): mention pre-existing formal benchmarks in addition to the informal benchmarks (MATH, NaturalProofs) --- e.g. TPTP, MPTP, 
% see: https://www21.in.tum.de/~wimmers/proofground/
% see https://do.proof.in.tum.de/
% also cite the CADE competition for ATPs http://www.tptp.org/CASC/
% also cite the DeepMind math dataset https://github.com/deepmind/mathematics_dataset
% also cite Tony Wu's work on INT
% cite 100 theorems list

\subsection*{Benchmarks}
In the closely related field of (first-order) \emph{automated theorem proving} (ATP), the TPTP~\citep{Sut17TPTP} benchmark is a library of test problems in a unified format for ATP systems. In interactive theorem proving, the "Freek 100"~\citep{wiedijk2008formalizing} tracks progress across various interactive theorem provers on a list of 100 mathematical theorems. \citet{wu2020int} built a simplified formal proof environment INT with an associated synthetic inequality benchmark. Competitions and communal challenges have also spurred development in formal theorem proving. The CADE ATP System Competition (CASC)~\citep{Sut16} is a competition that evaluates the performance of first-order automated theorem proving systems. Proof Ground~\citep{proofground}, part of the ITP conference, is an interactive proving contest (for humans) that supports Coq, Isabelle, and Lean, which focuses on evaluating the formalization effort of proof to given problems within limited time. Finally, the IMO Grand Challenge~\citep{imograndchallenge}, a proposal from researchers working on the interactive proof assistant Lean, aims to build a system capable of solving IMO problems in the formal-to-formal format.

Due to its convenient framing as a natural language processing task, the domain of informal mathematical reasoning has received more attention than the formal one. MATH~\citep{hendrycks2021measuring} is a mathematics benchmark comprising 12,500 statements in natural language where exercises are classified into 5 levels of difficulty across various domains. Each exercise is combined with a detailed step-by-step proof in natural language. Scaling state-of-the-art models shows little amelioration on MATH, which requires advanced mathematical reasoning capabilities. \textsf{miniF2F} includes a number of formalized statements from MATH. NaturalProofs~\citep{welleck2021naturalproofs} is another benchmark of natural proof in mathematics , containing 32k theorem statements and proofs. It essentially contains the proofs in ProofWiki and other resources. While MATH is more oriented towards mathematics exercises, NaturalProofs is focused on proofs of general mathematics theorems. \citet{saxton2019analysing} built a mathematics dataset with $2\times10^6$ training data and $10^4$ test data, presented in a question-answering format where each statement is paired with a question written in natural language and a direct answer without proof.

\subsection*{Neural theorem proving}
HOList~\citep{bansal2019holist,bansal2019learning,paliwal2020graph} provides an environment as well as a benchmark for HOL Light. They also proposes various deep reinforcement learning approaches for theorem proving and report a pass rate of $59.91\%$ on their benchmark. \citet{yang2019learning} built CoqGym, a large-scale dataset, which comes also with a learning environment, of 71k human-written proofs in Coq proof assistant. They report a $30.0\%$ pass rate on the held-out test theorems in CoqGym. \citet{polu2020generative} applied a decoder-only transformer similar to GPT-3~\citep{brown2020language} to proof steps prediction in Metamath combined with a log-probability based proof search. They also proposed a methodology to train a value function to further guide proof search, achieving a $56.22\%$ pass rate on the held-out test set. Large language models were applied to Lean by \citet{han2021proof}. They created an environment around the Lean prover targeted to machine learning and propose a dataset extracted from low level proof artifacts that is shown to boost performance when used as a self-supervised co-training objective. They report a $48.4\%$ pass rate on held-out test statements from \texttt{mathlib}, Lean's mathematical library \citep{DBLP:conf/cpp/X20}.

\section{miniF2F benchmark}

\begin{table}
\centering
\caption{Number of statements and their provenance in \textsf{miniF2F} \texttt{v1}}
\begin{tabular}{c|c|c|cc}
\multicolumn{1}{l}{}                                  & \multicolumn{1}{l}{}           & \multicolumn{1}{l}{} & \multicolumn{1}{l}{}         & \multicolumn{1}{l}{}      \\
\multicolumn{1}{c}{}                         & \multicolumn{1}{c}{}           & \multicolumn{1}{c}{} & \multicolumn{1}{l}{Test Set} & Validation Set  \\ 
\hline\hline
\multicolumn{3}{c|}{TOTAL}                                                                           & 244                          & 244             \\ 
\hline
\multicolumn{3}{c|}{\textbf{IMO}}                                                                             & 20                           & 20              \\
\multicolumn{3}{c|}{\textbf{AIME}}                                                                            & 15                           & 15              \\
\multicolumn{3}{c|}{\textbf{AMC}}                                                                             & 45                           & 45              \\ 
\cline{1-3}
\multirow{10}{*}{\textbf{MATH}}                       & \multirow{5}{*}{Algebra}       & Level 5              & 14                           & 14              \\
                                             &                                & Level 4              & 14                           & 14              \\
                                             &                                & Level 3              & 14                           & 14              \\
                                             &                                & Level 2              & 14                           & 14              \\
                                             &                                & Level 1              & 14                           & 14              \\ 
\cline{2-3}
                                             & \multirow{5}{*}{Number Theory} & Level 5              & 16                           & 16              \\
                                             &                                & Level 4              & 11                           & 11              \\
                                             &                                & Level 3              & 11                           & 11              \\
                                             &                                & Level 2              & 11                           & 11              \\
                                             &                                & Level 1              & 11                           & 11              \\ 
\cline{1-3}
\multicolumn{1}{l|}{\multirow{3}{*}{\textbf{CUSTOM}}} & \multicolumn{2}{c|}{Algebra}                          & 18                           & 18              \\
\multicolumn{1}{l|}{}                        & \multicolumn{2}{c|}{Number Theory}                    & 8                            & 8               \\
\multicolumn{1}{l|}{}                        & \multicolumn{2}{c|}{Induction}                        & 8                            & 8               \\
\hline
\end{tabular}
\label{tab: mf2f}
\end{table}

\textsf{miniF2F} is a dataset of manually formalized statements of Olympiad type problems, aligned in Lean, Metamath, and Isabelle (partial at the time of writing), providing a cross-platform benchmark for formal mathematical reasoning. Olympiad type problems are of particular interest to compare automated provers across different formal systems as the theories required to solve them are well identified and they generally do not require the definition of new mathematical concepts (a capability that remains beyond the current neural theorem proving state of the art).
 
The formalized statements in \textsf{miniF2F} are drawn from multiple sources, ranging from high school and undergraduate level exercises to Olympiad problems. \textsf{miniF2F} also covers different sub-subjects in mathematics as well as proof strategies, focusing on the types of exercises whose statements are expressible in most formal systems. This leads to a systemic focus on algebra, number theory and inequalities because, for example, geometry and combinatorial problems are generally challenging to formalize due to only nascent efforts in these areas in most formal systems. The statements in \textsf{miniF2F} are all manually formalized and selected to cover a variety of difficulty levels for both humans and machines. Formal proofs for these statements are optionally attached.

\textsf{miniF2F} draws from AIME, AMC, IMO problems as well as problems from the MATH~\citep{hendrycks2021measuring} informal dataset. Formalizing problems from the MATH dataset serves two purposes. First, problems in MATH are segmented by difficulty level (from $1$ to $5$), randomly selecting a subset from each of these difficulty levels allows \textsf{miniF2F} to cover a wider range of difficulty. Second, it provides the community an opportunity to compare capabilities of formal automated prover to their informal counter-parts as discussed in later sections.

\textsf{miniF2F} comprises a test set and a validation set, which are a stratified random split from the statements we formalized such that each set equally covers each problem type and difficulty (when available). Table \ref{tab: mf2f} shows a detailed distribution of these statements.

\paragraph*{Versioning}
\textsf{miniF2F} is an evolving effort and new statements will continuously be added. Periodically, we will freeze versions of the benchmark. The current version of the benchmark is \texttt{v1}\footnote{
\url{https://github.com/openai/miniF2F/tree/v1}
% masked for reviewers
} and results in this paper are reported using this version. \texttt{v1} comprises $244$ test and $244$ valid statements. The set of statements of each version is guaranteed to remain stable, only allowing fixes in case errors are later discovered.

\paragraph*{Rules of engagement and License} \textsf{miniF2F} is meant to serve as a shared resource for research groups working on applying deep learning to formal theorem proving. There is no formal process to submit evaluation results and researchers are simply invited to cite \textsf{miniF2F} indicating the version used in their evaluations. We also encourage them to contribute proofs found by their approaches back to the benchmark. The parts of the benchmark associated with each theorem prover (Metamath, Lean, Isabelle) are meant to be licensed in a way that is aligned with the licensing usage associated with the theorem prover's main library. As a result, the Metamath version of the benchmark is released under the MIT License, while the Lean and Isabelle versions are released under the Apache License.

\paragraph*{Formalization effort and challenges}
We found that, for trained practitioners (but not necessarily experts, including students recently introduced to formal systems), formalizing a statement takes about 15 minutes on average, and reviewing a formalized statement, about half of that on average. Note that not all exercises are directly or naturally formalizable. In particular, multi-choice questions, word problems, and exercises that require to explicit a witness or a set as part of the answer present interesting challenges:
\begin{description}
    \item \textit{multi-choice questions}\footnote{
        Example: \texttt{amc12a\_2020\_p10} in \url{https://github.com/openai/miniF2F/blob/main/lean/src/test.lean}
    } these problems are generally straightforwardly formalizable by reformulating the statement using the right answer only, and could be made “fair” in a competitive setup by formalizing all possible choices and running automated provers on all of them, attributing points only if a proof of the correct answer is provided.
    \item \textit{word problems}\footnote{
        Example: \texttt{mathd\_algebra\_398} in
        \url{https://github.com/openai/miniF2F/blob/main/lean/src/test.lean}
    } where significant information is presented in natural language generally require non-trivial efforts to be formalized. We generally formalized them by explicitly modeling the mathematics concepts and expression presented in natural language while attempting as best as possible to preserve the mathematical difficulty of the original problem. Sometime the formalization work is most of the difficulty associated with the original question; in such cases we would discard the problem entirely.
    \item \textit{problems that require to explicit a set or witness}\footnote{
        Example: \texttt{imo\_1997\_p5} in
        \url{https://github.com/openai/miniF2F/blob/main/lean/src/test.lean}
    } (e.g. find all ... such that ...) are not directly formalizable. The best approximation we relied on for these was to formalize the statement with the witness or answer provided, turning such exercises into the generation of a proof that the answer is correct, and if needed, that it is the unique one--which is, at times, a much easier exercise. A non negligible portion of IMO problems are as such, which we foresee could become a challenge in the future, to fairly compare humans to automated proving systems in a competitive setup.
\end{description}

\paragraph*{Porting effort}
In addition to Metamath, Lean, Isabelle (work in progress) and HOL Light (work in progress), we are eager to extend the coverage of \textsf{miniF2F} to Coq, and will welcome any effort in that direction or to extend \textsf{miniF2F} to further systems.

\section{Experiments}
In this section, in order to study baseline performances associated with existing systems, we report pass rates achieved by GPT-$f$~\citep{polu2020generative} applied to Metamath, GPT-$f$/PACT~\citep{polu2020generative,han2021proof} applied to Lean as well as a baseline prover implemented in Lean denoted as the \texttt{tidy} baseline. Pass rates are reported as $\text{Pass}@N$ where $N$ is the number of proof search attempts per statement. $\text{Pass}@N$ is computed by running more attempts per statement, averaged to get an unbiased, low-variance estimate.

\subsection{Metamath}

Metamath is powered by a meta logic system based on a single substitution rule. It's characterized by its simplicity which makes it convenient to study machine learning. Proofs in Metamath are, as a consequence of the low-level proofsteps, much longer than in other systems as there is no assistance from high-level tactics. Proofs which are trivial in other systems (e.g. n-digit addition or simple ring arithmetic transformations) can be quite tedious in Metamath. The absence of tactics is both a benefit, as the models sees and learns on everything, and a challenge, as proofs of even simple exercises require hundreds of proofsteps.

\subsubsection{GPT-f}

We report the pass rate of GPT-$f$ applied to Metamath as described in \citet{polu2020generative}. We use a model with 700m learnable parameters. The model is trained on an updated dump of the set.mm library (but similar synthetic datasets), using the log-probability based search as reported in Table 8 of the GPT-$f$ paper~\citep{polu2020generative}.

The model achieves a $\text{Pass}@1$ of $1.3\%$ and a $\text{Pass}@8$ of $1.6\%$ on \textsf{miniF2F}-test. As expected, these numbers are quite low due to the length of typical proofs for even simple math exercises. The average proof length is also reported in Table \ref{tab:result}.

\subsection{Lean}

In comparison to Metamath, Lean benefits from a large number of powerful tactics to assist formalization efforts. Typical Lean proofs are much shorter than Metamath's. This is also a formal system of interest as it has received a lot of attention from the mathematical community as recent theories have successfully been formalized in Lean (Perfectoid Spaces~\citep{perfectoidspaces}, Liquid Tensor experiment~\citep{liquidtensor}).

Lean is also associated with the IMO Grand Challenge~\citep{imograndchallenge} which aims to organize a formal-to-formal challenge during the upcoming IMO competitions.

\begin{figure}
    \centering
    \includegraphics[width=\linewidth]{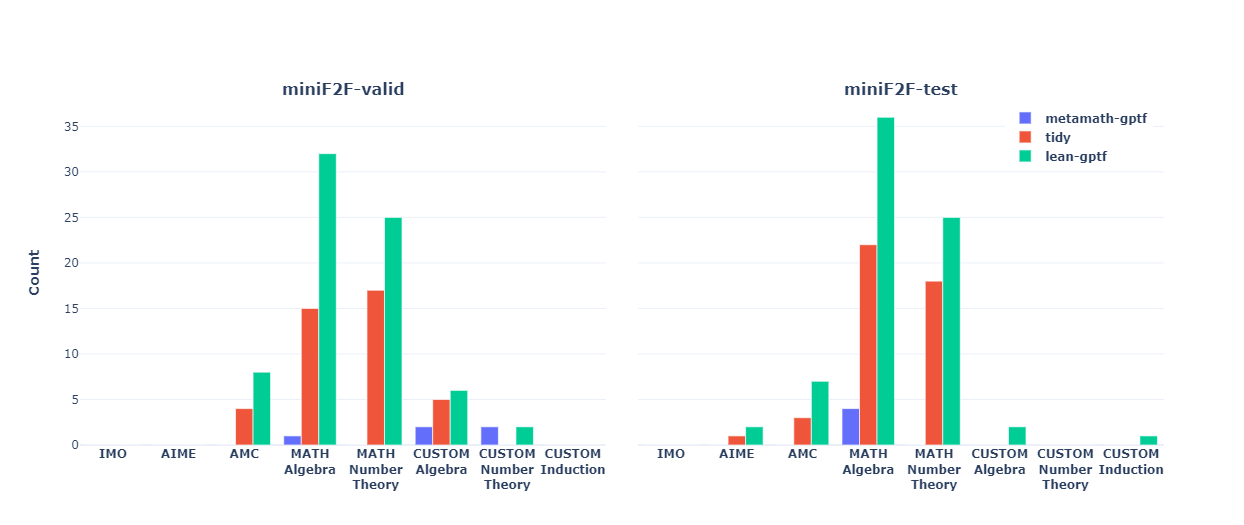}
    \caption{Counts of successfully proved statements in \textsf{miniF2F}. Green bar: results from Lean GPT-f. Red bar: best result from the \texttt{tidy} baseline. Blue bar: results from Metamath GPT-f.}
    \label{fig:barplot}
\end{figure}

\subsubsection{\texttt{tidy} baseline}
\label{subsection:tidy}

We use the generic best-first search algorithm presented in PACT~\citep{han2021proof}. The algorithm works as follows: Given a list of tactics $L$ with priority, we maintain a priority queue $Q$ of tactic states whose priority is given by the priority of the last applied tactic in $L$ that led to it. While $Q$ is not empty, we pop the top tactic state $t$ from $Q$. We iterate through $L$ and apply each tactic to $t$. If no error is raised, we capture the returned tactic states from Lean and insert them back into $Q$.

We use the same terminology as in PACT~\citep{han2021proof}: maximum queue size $\omega_{max}$, depth limit $d_{max}$. We also enforce a budget of $i_{max}$ iterations of the outer loop. When $Q$'s size reach $q_{max}$, all the tactic states to be inserted are discarded. We do not expand the next tactic state when the depth is beyond $d_{max}$. This loop is run until a proof is found or the iterations budget is exhausted.

For consistency checking, we run the \texttt{tidy} baseline under the same settings and on the same test set as in PACT~\citep{han2021proof} except that we don't set a global timeout. Our implementation achieved a 10.5\% pass rate on mathlib's test split. This result is comparable to the reported 9.9\% in PACT given the waived global timeout.

In addition to the curated list of tactics $L$ used in PACT~\citep{han2021proof}, we added 4 high-level tactics $HL=$\texttt{[nlinarith, linarith, ring\_nf, norm\_num]} to $L$ with higher priorities than the others. We report our pass rate on \textsf{miniF2F} in Table \ref{tab:tidy}.

\begin{table}[ht]
\centering
\caption{The table shows the number of solved statement in \textsf{miniF2F} when running the \texttt{tidy} baseline with different values of $i_{max}$ as well Lean's built-in \texttt{tidy} tactic. All \texttt{tidy} baseline experiments are run with $\omega_{max}=128$, $d_{max} = 8$ using $L + HL$. Despite the \texttt{tidy} baseline being deterministic, it is still subject to per-tactic application timeouts, explaining the number $43$ reported on \textsf{miniF2F}-test for $i_{max} = 32$.}
\label{tab:tidy}
\begin{tabular}{lcc}
                                                             & \multicolumn{1}{l}{}              & \multicolumn{1}{l}{}      \\
\multicolumn{1}{c}{parameters}                               & \multicolumn{1}{l}{miniF2F-valid} & miniF2F-test              \\ 
\hline\hline
\begin{tabular}[c]{@{}l@{}}Lean's tidy tactic\\\end{tabular} & 12~/ 244          & 13~/ 244  \\ 
\hline
$i_{max} = 1$                                                & 21 / 244                          & 23~/ 244  \\
$i_{max} = 2$                                                & 31~/ 244          & 29~/ 244  \\
$i_{max} = 4$                                                & 38~/ 244          & 41~/ 244  \\
$i_{max} = 8$                                                & 41 / 244                          & 44~/ 244  \\
$i_{max} = 16$                                               & 41~/ 244          & 44~/ 244  \\
$i_{max} = 32$                                               & 41~/ 244          & 43~/ 244  \\
$i_{max} = 64$                                               & 41~/ 244          & 44~/ 244  \\
$i_{max} = 128$                                              & 41~/ 244          & 44~/ 244 
\end{tabular}
\end{table}

\subsubsection{GPT-f/PACT}

We report the pass rate of GPT-$f$/PACT as described in~\citet{han2021proof}. We use a model with 700M learnable parameters. The model is trained on an updated dump\footnote{\url{https://github.com/jasonrute/lean_proof_recording/commit/8499f10c2e10dd533152070ed933c4f0b21ecdc0}}\footnote{\url{https://github.com/jesse-michael-han/lean-step-public/commit/a2b83c237bfe4d6f1c48bb48bc0769b5940e614a}} of the mathlib library using the PACT methodology denoted in the paper as \verb$mix2 > mix1 + tactic$ in Figure~6.

The model achieves a $\text{Pass}@1$ of $24.6\%$ and a $\text{Pass}@8$ of $29.2\%$ on \textsf{miniF2F}-test. The average proof length is also reported in Table \ref{tab:result}.

\begin{table}[ht]
\centering
\caption{Baseline performance on Metamath and Lean. All proof searches are provided with a $128$ expansions budget. GPT-$f$ attempts $e=16$ tactics per expansion while the \texttt{tidy} baseline attempts $e=17$ tactics per expansion ($L + HL$, see section \ref{subsection:tidy}). Reported proof lengths are averages over all the proofs found in each run. Note that the \texttt{tidy} baseline being deterministic, there is no point attempting a proof search more than once.}
\label{tab:result}
\begin{tabular}{cccccccccc}
\multicolumn{1}{l}{}                                                    & \multicolumn{1}{l}{}   & \multicolumn{1}{l}{} & \multicolumn{3}{l}{}                                                                                                                                  &                      & \multicolumn{3}{l}{}                                                                                                               \\
                                                                        &                        &                      & \multicolumn{3}{c}{miniF2F-valid}                                                                                                                     & \multicolumn{1}{c}{} & \multicolumn{3}{c}{miniF2F-test}                                                                                                   \\ 
\hhline{==~===~===}
\multirow{2}{*}{\begin{tabular}[c]{@{}c@{}}Formal\\System\end{tabular}} & \multirow{2}{*}{Model} & \multirow{2}{*}{}    & \multicolumn{1}{l}{\multirow{2}{*}{\begin{tabular}[c]{@{}c@{}}Proof \\Length\end{tabular}}} & \multicolumn{2}{c}{Pass rate}                           & \multicolumn{1}{c}{} & \multirow{2}{*}{\begin{tabular}[c]{@{}c@{}}Proof \\Length\end{tabular}} & \multicolumn{2}{c}{Pass rate}                            \\
                                                                        &                        &                      & \multicolumn{1}{l}{}                                                                        & \multicolumn{1}{l}{Pass@1} & \multicolumn{1}{l}{Pass@8} &                      &                                                                         & \multicolumn{1}{l}{Pass@1} & \multicolumn{1}{l}{Pass@8}  \\ 
\hhline{==~===~===}
Metamath                                                                & GPT-$f$                  &                      & 16.2                                                                                        & 1.0\%                      & 2.0\%                      &                      & 20.3                                                                    & 1.3\%                      & 1.6\%                       \\
Lean                                                                    & tidy                   &                      & 1.7                                                                                         & 16.8\%                     & -                          & \multicolumn{1}{c}{} & 1.8                                                                     & 18.0\%                     & -                           \\
Lean                                                                    & GPT-$f$                  &                      & 2.6                                                                                         & 23.9\%                     & 29.3\%                     &                      & 2.5                                                                     & 24.6\%                     & 29.2\%                     
\end{tabular}
\end{table}

\subsection{Discussion}

\subsubsection{Access to high-level tactics}
One goal of \textsf{miniF2F} is to study the comparison of performance across formal systems. In this section we reported the performance of the same methodology (GPT-$f$~\citep{polu2020generative}) applied to both Lean and Metamath. Both models are pre-trained on WebMath~\citep{polu2020generative} and respectively trained on datasets extracted from Lean~\citep{han2021proof} and Metamath~\citep{polu2020generative}. The overall compute deployed at training is comparable in both setup and exactly equivalent at test-time, yet the achieved performance appears drastically superior when applied to Lean. We hypothesize that this is mainly explained by the model's access to high-level tactics when applied to Lean, enabling the model to learn how to guide Lean's automation in an effective way.

An example of this high-level guidance behavior is well exemplified by the following proof of the statement \verb+algebra_sqineq_2unitcircatblt1+ where the model heavily relies on Lean's \verb_nlinarith_ solver but provides it with essential premises to successfully guide the search.

\begin{verbatim}
theorem algebra_sqineq_2unitcircatblt1
  (a b : ℝ)
  (h₀ : a^2 + b^2 = 2) :
  a * b ≤ 1 :=
begin
  nlinarith [sq_nonneg a,sq_nonneg b,sq_nonneg (a - b)]
end
\end{verbatim} 

(The statement above (\verb+algebra_sqineq_2unitcircatblt1+) requires to prove the assertion $\forall a, b \in \mathbb{R}, a^2 + b^2 = 2 \rightarrow a \cdot b \leq 1$).

In Metamath, GPT-$f$ fails to find a proof as it requires a very large number of steps to appropriately rewrite the goal in a way that is amenable to the use of set.mm's existing theorems. The \texttt{tidy} baseline also fails to find a proof of that statement as \verb_nlinarith_ is not capable of solving the goal without being passed extraneous premises.

These results motivate the use of neural theorem proving with formal systems that expose powerful high level tactics and also suggest the potential of a closer collaboration between formal systems and machine learning practitioners. It also motivates the use of generative models in that setup as the arguments required by high-level tactics to succeed on non trivial problems generally do not exist in the context of the statement and therefore have to be generated ex-nihilo.

\subsubsection{Comparison of informal and formal setups}
The use of formal systems for neural theorem proving is often motivated by the role of the formal system as a verifier, enabling more advanced neural search strategies than possible in a fully informal setup where the generation of a model can't be verified automatically, as well as the access to powerful tactics. Our formalization of a subset of the MATH~\citep{hendrycks2021measuring} informal dataset provides an interesting approximate quantification of the benefit of having access to a formal system in the context of neural theorem proving. Approximate, because we only formalized a small subset of the MATH statements, but nonetheless useful since we drew uniformly from the 5 difficulty levels.

In~\citet{hendrycks2021measuring}, the performance of GPT-3 (which is a larger model than the GPT-f model studied here) is reported to be 6.0\% in the algebra category and 3.9\% in the number theory category. GPT-$f$ applied to Lean by comparison achieves 51.4\% in the algebra category and 41.7\% in the number theory category. It is also worthwhile to note that the \texttt{tidy} baseline also highly outperforms (31.4\% in algebra and 30.0\% in number theory) GPT-3 in an informal setup demonstrating the benefit of proof automation alone.

\subsubsection{Limitation}
With \textsf{miniF2F} being cross-system as the goal, types of problems that are less expressible in certain systems such as geometry and combinatorial problems are less covered. The shift of distribution of problem types may result in skewing the research direction of models when benchmarking on \textsf{miniF2F}. Directionally we aim to fix it and extend the coverage of \textsf{miniF2F} as we grow the benchmark. However, works and efforts on the corresponding library of other systems are required as well.

\section{Conclusion}
We presented \textsf{miniF2F}, a dataset of formal Olympiad-level mathematics problem statements, meant to serve as an initial effort towards cross-system benchmarking of neural mathematical reasoning capabilities in formal environments. We reported the performance of the neural theorem prover GPT-$f$~\citep{polu2020generative} on both the Lean and Metamath parts of \textsf{miniF2F} as well as the performance of our non-neural \texttt{tidy} baseline applied to Lean. Then, we discussed these baselines and put them in perspective with previously reported comparable results in informal environments~\citep{hendrycks2021measuring}.

Finally, we hope that \textsf{miniF2F} will prove to be useful to the scientific community working on neural theorem proving and spur advances in this domain.

\subsubsection*{Acknowledgments}
We are grateful to Wenda Li and Xavier Martinet for contributing the Isabelle and HOL Light statements currently available in \textsf{miniF2F}, paving the way towards a full support of Isabelle and HOL Light, as well as their feedback and encouragement in the process. We thank Harri Edwards for his comments that greatly improved the manuscript.

\bibliography{iclr2022_conference}
\bibliographystyle{iclr2022_conference}

\newpage
\appendix
\appendix
\section{Example of statement in miniF2F}

\begin{table}[ht]
\centering
\caption{Problem 11 of 2000 AMC 12 is formalized with proof in different languages in \textsf{miniF2F}. The proof is optionally attached thus not part of the benchmark. The proof in Metamath is too long to be fully displayed.}
\begin{tabular}{|m{1.5cm}|m{10cm}|}
\multicolumn{1}{l}{} & \multicolumn{1}{l}{}\\ 
\hline
\centering Natural Language & Two non-zero real numbers, $a$ and $b,$ satisfy $ab = a - b$. Which of the following is a possible value of $\frac {a}{b} + \frac {b}{a} - ab$? (A) -2 (B) $\frac{-1}{2}$ (C) $\frac{1}{3}$ (D) $\frac{1}{2}$ (E) 2 \\ 
\hhline{|==|}
\centering Metamath         & 
\texttt{\small\makecell[l]{\$\{ \\
  amc12-2000-p11.0 \$e |- ( ph -> A e. RR ) \$.\\
  amc12-2000-p11.1 \$e |- ( ph -> B e. RR ) \$.\\
  amc12-2000-p11.2 \$e |- ( ph -> A =/= 0 ) \$.\\
  amc12-2000-p11.3 \$e |- ( ph -> B =/= 0 ) \$.\\
  amc12-2000-p11.4 \$e |- ( ph -> ( A x. B ) =
  \\
  \quad ( A - B ) ) \$.\\
  amc12-2000-p11 \$p |- ( ph -> ( ( ( A / B ) + \\
  \quad ( B / A ) ) - ( A x. B ) ) = 2 ) \\
\$=\\
( cdiv co caddc cmul cmin c2 cexp eqcomd ... \$. \\
\$\} \\ }}

\\ \hline
\centering Lean             & 
\texttt{\makecell[l]{
theorem amc12\_2000\_p11 \\
\quad (a b : $\mathbb{R}$) \\
\quad  (h₀ : a ≠ 0 ∧ b ≠ 0) \\
\quad  (h₁ : a * b = a - b) : \\
\quad  a / b + b / a - a * b = 2 := \\
begin \\
\quad  field\_simp [h₀.1, h₀.2], \\
\quad  simp only [h₁, mul\_comm, mul\_sub], \\
\quad  ring,\\
end \\}}
\\ \hline
\centering Isabelle         &  
\texttt{\makecell[l]{
theorem amc12\_2000\_p11:\\
\quad fixes a b::real \\
\quad  assumes "a $\setminus $$<$noteq$>$ 0" "b $\setminus $$<$noteq$>$ 0" \\
\quad \quad       and "a * b = a - b" \\ 
\quad \quad     shows "a / b + b / a - a * b = 2" \\
\quad   using assms \\ 
\quad   by (smt (verit, ccfv\_threshold) \\
\quad \quad diff\_divide\_distrib \\
\quad \quad div\_self divide\_divide\_times\_eq \\
\quad \quad eq\_divide\_imp nonzero\_mult\_div\_cancel\_left) \\
end \\}}
\\ \hline
\end{tabular}
\label{tab:ex_minif2f}
\end{table}

\newpage
\section{Performance by difficulty on statements formalized from MATH dataset}

The MATH dataset assigns a difficulty ranging from $1$ to $5$ to each of its problem. Tables \ref{tab: algebraMATH} and \ref{tab: numbertheoryMATH} report the number of proved statement split by difficulty level on the algebra and number theory categories.

% \begin{table}
% \centering
% \caption{Counts of successfully proved statements in MATH-Algebra, subset of \textsf{miniF2F}. This table corresponds to the bar ``MATH Algebra" in Figure \ref{fig:barplot} when breaking down the difficulties.}
% \label{tab: mf2f}
% \begin{tabular}{cccclccc}
%               & \multicolumn{3}{c}{miniF2F - Valid}                                             &  & \multicolumn{3}{c}{miniF2F - Test}     \\
%               & \begin{tabular}[c]{@{}c@{}}Metamath/GPT-$f$\\\end{tabular} & Lean/tidy & Lean/GPT-$f$ &  & Metamath/GPT-$f$ & Lean/tidy & Lean/GPT-$f$  \\ 
% \hhline{====~===}
% Level 1 & 1                                                       & 6         & 9         &  & 2             & 6         & 8          \\
% Level 2 & 0                                                       & 4         & 7         &  & 0             & 4         & 7          \\
% Level 3 & 0                                                       & 2         & 8         &  & 1             & 7         & 10         \\
% Level 4 & 0                                                       & 2         & 6         &  & 0             & 3         & 7          \\
% Level 5 & 0                                                       & 1         & 2         &  & 1             & 1         & 3          \\
%               &                                                         &           &           &  &               &           &           
% \end{tabular}
% \end{table}

\begin{table}[ht]
\centering
\caption{Counts of successfully proved statements formalized from MATH-Algebra in \textsf{miniF2F} \texttt{v1} split by difficulty. This table corresponds to ``MATH Algebra" in Figure \ref{fig:barplot}.}
\label{tab: algebraMATH}
\begin{tabular}{cccccccccccc}
\multicolumn{1}{l}{} & \multicolumn{5}{l}{}              & \multicolumn{1}{l}{} & \multicolumn{5}{l}{}              \\
                     & \multicolumn{5}{c}{miniF2F-valid} &                      & \multicolumn{5}{c}{miniF2F-test}  \\
Difficulty Level     & 1 & 2 & 3 & 4 & 5                 &                      & 1 & 2 & 3  & 4 & 5                \\ 
\hhline{======~=====}
Metamath/GPT-$f$     & 1 & 0 & 0 & 0 & 0                 &                      & 2 & 0 & 1  & 0 & 1                \\
Lean/tidy            & 6 & 4 & 2 & 2 & 1                 &                      & 6 & 4 & 7  & 3 & 1                \\
Lean/GPT-$f$         & 9 & 7 & 8 & 6 & 2                 &                      & 8 & 7 & 10 & 7 & 3               
\end{tabular}
\end{table}

\begin{table}[ht]
\centering
\caption{Counts of successfully proved statements formalized from MATH-Number theory in \textsf{miniF2F} \texttt{v1} split by difficulty. This table corresponds to ``MATH Number Theory" in Figure \ref{fig:barplot}.}
\label{tab: numbertheoryMATH}
\begin{tabular}{cccccccccccc}
\multicolumn{1}{l}{} & \multicolumn{5}{l}{}              & \multicolumn{1}{l}{} & \multicolumn{5}{l}{}              \\
                     & \multicolumn{5}{c}{miniF2F-valid} &                      & \multicolumn{5}{c}{miniF2F-test}  \\
Difficulty Level     & 1 & 2 & 3 & 4 & 5                 &                      & 1  & 2 & 3 & 4 & 5                \\ 
\hhline{======~=====}
Metamath/GPT-$f$     & 0 & 0 & 0 & 0 & 0                 &                      & 0  & 0 & 0 & 0 & 0                \\
Lean/tidy            & 8 & 3 & 2 & 2 & 2                 &                      & 7  & 4 & 3 & 2 & 2                \\
Lean/GPT-$f$         & 9 & 5 & 5 & 4 & 2                 &                      & 10 & 5 & 5 & 3 & 2               
\end{tabular}
\end{table}

More broadly, Lean GPT-$f$ is capable of solving any problem that the \texttt{tidy} baseline or Metamath GPT-$f$ can solve in \textsf{MiniF2F}. Qualitatively, the problems on which it fail either require multiple non-trivial reasoning steps (outside a few exceptions, problems requiring more than 2 non-trivial steps of mathematical reasoning are generally out of reach of these baselines) or require a cut introduction that is hard to generate, such as generating a non trivial witness.

% The performance across difficulties on statements sourced from MATH dataset(reported in Table \ref{tab: algebraMATH} and \ref{tab: numbertheoryMATH}) shows that the count of successfully proved statements decreases as the difficulty increases, and again, the performance gap between Metamath and Lean system.

% Due to the access of high-level tactic, both the \texttt{tidy} baseline and GPT-$f$/PACT in Lean's setup outperforms greatly the Metamath's. This advantage is even more magnified in the subset of number theory, resulting in the 0 successfully proved statement across all difficulties by GPT-$f$ in Metamath's setup: Low-difficulty statements formalized  from  MATH-Number theory in \textsf{miniF2F} are mostly plug-and-chug in Lean, e.g. showing that the greatest-common-divisor of 180 and 168 is 12(statement \texttt{mathd\_numbertheory\_188}, classified as Level 2 in MATH) can be solved by a single application of \texttt{norm\_num}, a tactic designed for evaluating arithmetic expressions. While in Metamath, this requires recursive applications of Euclid's algorithm and thus far more reasoning steps. 

\end{document}